\documentclass{article} 
\usepackage{iclr2025_conference,times}

\usepackage{amsmath,amsfonts,bm}

\def\eqref#1{equation~\ref{#1}}

\def\1{\bm{1}}

\DeclareMathAlphabet{\mathsfit}{\encodingdefault}{\sfdefault}{m}{sl}
\SetMathAlphabet{\mathsfit}{bold}{\encodingdefault}{\sfdefault}{bx}{n}

\usepackage[colorlinks=true, urlcolor=cyan, citecolor=cyan]{hyperref}
\usepackage{url}
\usepackage{booktabs}
\usepackage{subcaption}
\usepackage{graphicx}
\usepackage{multicol}
\usepackage{multirow}
\usepackage{listings}
\usepackage[frozencache,cachedir=.]{minted}
\usepackage{enumitem}

\title{MatMamba: A Matryoshka State Space Model}

\author{Abhinav Shukla$^{\dagger}$
\And
Sai Vemprala$^{\dagger}$
\And
Aditya Kusupati$^{\diamond}$\thanks{AK is currently at Google DeepMind}
\And
Ashish Kapoor$^{\dagger}$
\AND
\@\vspace{-0.8cm}\\
$^{\dagger}$Scaled Foundations \enspace $^{\diamond}$University of Washington\vspace{2mm}\\
\texttt{\{abhinav,sai,ashish\}@scaledfoundations.ai, kusupati@cs.uw.edu}
}

\iclrfinalcopy 

\begin{document}

\maketitle

\begin{abstract}
State Space Models (SSMs) like Mamba2 are a promising alternative to Transformers, with faster theoretical training and inference times -- especially for long context lengths. Recent work on Matryoshka Representation Learning -- and its application to Transformer backbones in works like MatFormer --  showed how to introduce nested granularities of smaller submodels in one universal elastic model. In this work, we present MatMamba: a state space model which combines Matryoshka-style learning with Mamba2, by modifying the block to contain nested dimensions to enable joint training and adaptive inference. MatMamba allows for efficient and adaptive deployment across various model sizes. We train a single large MatMamba model and are able to get a number of smaller nested models for free -- while maintaining or improving upon the performance of a baseline smaller model trained from scratch. We train language and image models at a variety of parameter sizes from 35M to 1.4B. Our results on ImageNet and FineWeb show that MatMamba models scale comparably to Transformers, while having more efficient inference characteristics. This makes MatMamba a practically viable option for deploying large-scale models in an elastic way based on the available inference compute. Code and models are open sourced at \url{https://github.com/ScaledFoundations/MatMamba}
\end{abstract}

\section{Introduction}
Deep learning practitioners often train different sizes of the same kind of model to facilitate deployment in a variety of ranges of available inference compute. For example, the Llama 3.2~\citep{dubey2024llama} series has 1B, 3B, 11B, and 90B variations. These models are extremely powerful individually -- but due to independent training do not necessarily share the same metric space -- a property which can be extremely useful for inference applications like speculative decoding~\citep{leviathan2023fast}, hybrid cloud-edge inference, or just general input or compute adaptive processing. 
Moreover, because training these models is expensive, we typically see only a few chosen sizes trained. This is not desirable in situations where the deployment setup can optimally support an intermediate model (e.g. a 2B model), but has to settle for the less accurate 1B model instead.

Techniques like model compression and distillation aim to address these issues, but require additional training (for which data may not be available), and can sometimes drop accuracy~\citep{jaiswal2023compressing}. Thus, methods that offer adaptive inference out of the box at intermediate granularities are extremely useful. This has been explored for Transformers~\citep{kudugunta2023matformer, cai2024flextron} and ConvNets~\citep{yu2019universally, cai2019once}. The core focus of this work is to try to enable out of the box adaptive inference in a newer architecture: Mamba2~\citep{dao2024transformers}.

State Space Models like Mamba2~\citep{dao2024transformers} and a number of other related newer architectures (see Section \ref{sec:rw}) have shown tremendous potential as they try to improve on the efficiency of Transformers, while maintaining their potency as accurate and general sequence processing architectures. Mamba2 has comparable scaling properties to Transformers, while being significantly faster at longer context lengths.

In this work, we introduce MatMamba, a nested Matryoshka structure~\citep{kusupati2022matryoshka} within a Mamba2 block~\citep{dao2024transformers}. MatMamba enables the extraction of hundreds of nested submodels from the same set of weights, without requiring any additional training during deployment. MatMamba is a general-purpose sequence processing architecture that can be applied to any type of model (encoder/decoder), modality (language/vision/sound/actions), loss function, or learning algorithm compatible with a Transformer or Mamba2 layer.

\begin{figure}
    \centering
    \includegraphics[width=\linewidth]{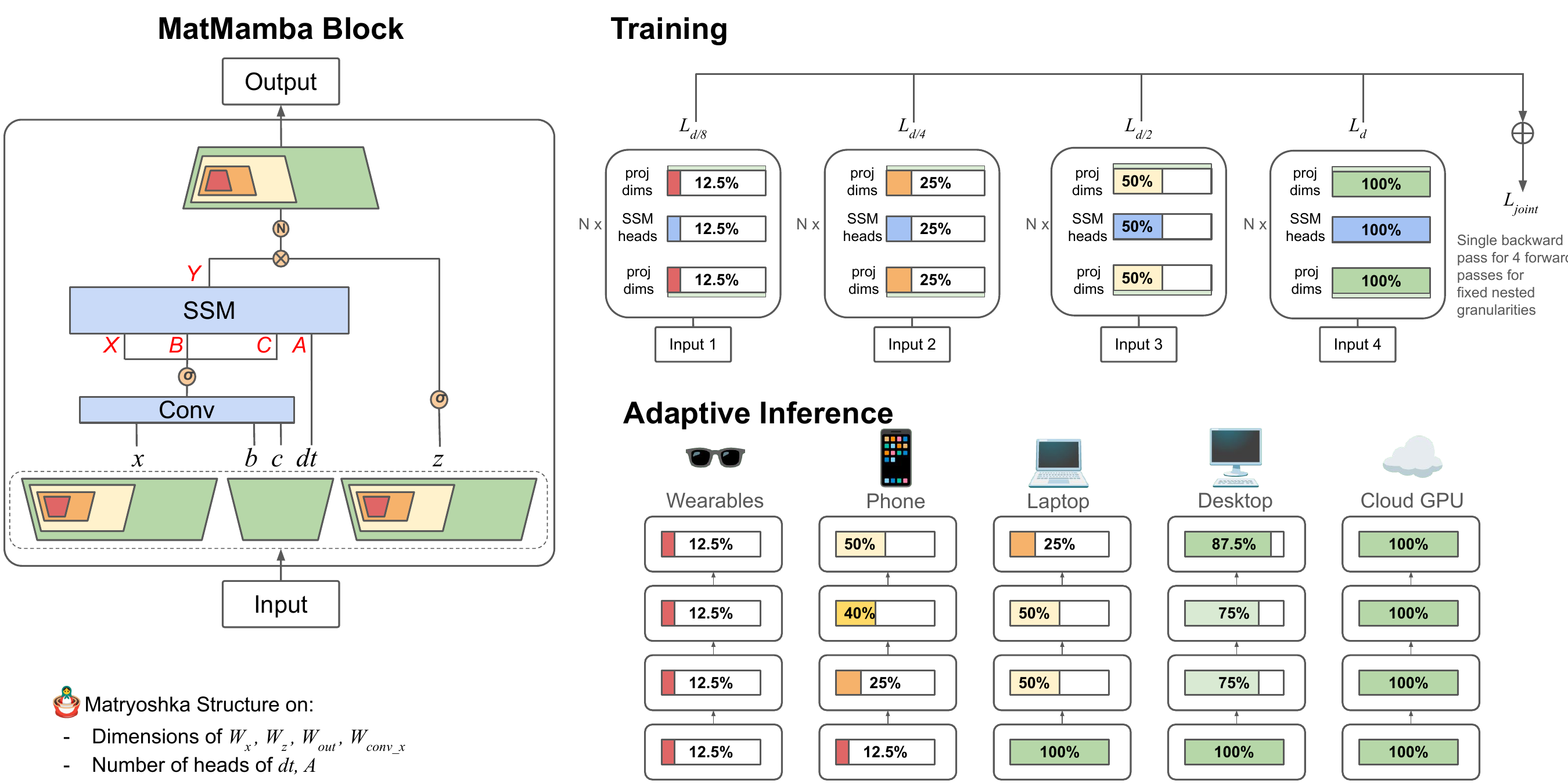}
    \caption{MatMamba introduces a nested Matryoshka~\citep{kusupati2022matryoshka} structure in a Mamba2~\citep{dao2024transformers} block. We jointly train a few chosen granularities to get a single model from which we can flexibly extract a large number of nested submodels for adaptive inference based on the available deployment compute. }
    \label{fig:matmamba-block}
\end{figure}

The philosophically closest work to MatMamba is MatFormer~\citep{kudugunta2023matformer} -- which imposes a nested structure on the FFN block in a Transformer layer. We use the same concept to impose a nested structure on any learnable parameter in a Mamba2 block that depends upon the hidden dimensionality of the block. Formally, a MatMamba block consists of a nested combination of $g$ Mamba2 blocks $M_i$, such that $M_1 \subset M_2 \subset ... \subset M_g$, where $M_i \subset M_j$ means that all the parameters of a sub-block $M_i$ are present in $M_j$. We train the model using $g$ forward passes with gradient accumulation followed by a single backward pass for parameter updates (see Figure \ref{fig:matmamba-block}).

By jointly training all $g$ granularities, the smallest sub-blocks are incentivized to represent the most important information, like in Matryoshka Representation Learning~\citep{kusupati2022matryoshka}. We can now use any of the $g$ nested sub-blocks $M_i$ flexibly. Additionally, we can flexibly slice the block along \textit{any} dimensionality (even beyond the $g$ explicitly optimized granularities). Using Mix'n'Match (Section \ref{sec:mixnmatch}), we can perform this operation over multiple layers at varying granularities to flexibly extract a combinatorially large number of models from the single larger model. We observe that these extracted models preserve the metric space of the larger model, and are accurate across a variety of tested tasks -- effectively allowing us to choose a tradeoff between model performance and compute.

We train MatMamba-based vision models (MatMamba-Vision), and find that: (a) MatMamba-Vision models scale as well as baseline Mamba2 based models at all $g=4$ granularities; (b) Using Mix'n'Match, we can flexibly extract submodels between the explicitly optimized granularities. The submodels span (and sometimes exceed) the pareto optimal accuracy-vs-compute curve; (c) MatMamba-Vision models are significantly faster at higher resolutions than ViTs, making them promising candidates for long-form and high resolution visual tasks, while enabling adaptive visual processing with the nested submodels (see Section \ref{sec:imageclassification}).

Furthermore, MatMamba-Vision models can act as elastic image encoders for adaptive image retrieval. We can encode visual datasets with the largest model, and because the smaller submodels share its metric space, we can use them as query encoders, needing drastically lower compute with minimal loss in accuracy (see Section \ref{sec:imageretrieval}).

We also train MatMamba-based decoder language models (MatMamba-LM) at various sizes from 130M-1.4B parameters, and at $g=4$ granularities. We make similar observations here too, that MatMamba-LM models scale as well as Mamba2 baselines with the same architecture for all nested granularities. We also observe interesting homogenous scaling behaviour between the nested granularities for different models (see Section \ref{sec:lm}).

Through MatMamba, for the first time, we bring together the adaptivity of Matryoshka-style learning and the efficiency of state space models (SSMs) like Mamba2~\citep{dao2024transformers}. 

\textbf{We make the following research contributions:}
\begin{enumerate}[leftmargin=*,align=left]
\item We introduce MatMamba, which imposes a nested Matryoshka structure on a Mamba2 state space model. We jointly optimize all nested granularities to train a single elastic model.
\item We show that MatMamba models scale as well as the baseline Mamba2 models for a variety of model sizes from 35M-1.4B parameters on language and vision tasks.
\item Using Mix'n'Match with MatMamba allows the flexible extraction of hundreds of submodels to perform adaptive inference. These submodels preserve the metric space of the original model.
\item MatMamba-Vision models are comparably accurate and significantly faster at higher resolutions than ViTs, making them well suited for long-form/high resolution and adaptive visual processing.
\end{enumerate}


\section{Related Work\label{sec:rw}}

The ever growing demand of AI models across various accuracy and resource constraints makes it infeasible to train a different model for each use case. Instead, these adaptive deployment needs are often solved through introducing elasticity in models~\citep{kusupati2024towards}. Work on slimmable networks~\citep{yu2018slimmable,yu2019universally} and once-for-all networks~\citep{cai2019once} brought the idea of training multiple submodels present within one universal model. Nested dropout~\citep{rippel2014learning} generalizes this idea to learn ordered representations which further extended to enable elasticity at each dense vector embeddings through Matryoshka Representation Learning (MRL)~\citep{kusupati2022matryoshka}. MRL simplifies the training process to induce elasticity with a small set of nested granularities (hence the name Matryoshka), exponentially separated in size, all optimized with the same target loss function as the full vector. MRL further smoothly interpolates to the granularities not seen during training, thus allowing for complete elasticity to extract sub-vectors based on the requirements.

Matryoshka information packing and learning has been widely adopted in bringing adaptivity not only in output space, but also in input~\citep{beyer2023flexivit} and model weights~\citep{kudugunta2023matformer,cai2024flextron,valipour2023sortednet}. MatFormer~\citep{kudugunta2023matformer} is a direct translation of MRL to every hidden activation vector of a MLP sub-block within a Transformer layer~\citep{vaswani2017attention}. MatFormer showed scaling trends similar to Transformer, while also providing the capability to adaptively extract submodels that fall on the accuracy-vs-compute pareto curve. More recent works~\citep{cai2024flextron,jain2024mixture} developed dynamic routing on top of the conditional computation enabled by MatFormer to realize performance gains in deployment. Further, matryoshka packing was also used for flexible tokenization~\citep{cai2024matryoshka,hu2024matryoshka} as well as diffusion models~\citep{gu2023matryoshka}.



Transformers~\citep{vaswani2017attention} have been fundamental sequence processing blocks in neural networks for the past few years. There has been a recent wave of work on efficient sequence processing architectures that aim to be faster and equally performant alternatives to Transformers.
Mamba~\citep{gu2023mamba} and Mamba2~\citep{dao2024transformers} are the most relevant to this work, with other very closely related works like Linear Attention~\citep{katharopoulos2020transformers}, Test-time training~\citep{sun2024learning}, RWKV~\citep{peng2023rwkv}, Griffin~\citep{de2024griffin}, Jamba~\citep{lieber2024jamba}, xLSTM~\citep{beck2024xlstm}, HGRN2~\citep{qin2024hgrn2}, RetNet~\citep{sun2023retentive}, RecurrentGemma~\citep{botev2024recurrentgemma}.
\citet{waleffe2024empirical} present a detailed study of how to train large-scale Mamba-based language models. Works like MambaVision~\citep{hatamizadeh2024mambavision}, MambaND~\citep{li2024mamba}, Vision Mamba~\citep{zhu2024vision}, 
VideoMamba~\citep{li2024videomamba}, and Sonic~\citep{Sonic} have all shown how a Mamba layer can process visual data and other modalities. \citet{liu2024vision} present a detailed survey of Mamba-based vision models.
\section{MatMamba} 

\subsection{Mamba2 Preliminaries}

MatMamba is based on Mamba2. We make simple modifications to the Mamba2 block to impose the Matryoshka structure. A detailed description of the internals of Mamba2 can be found in the original paper \cite{dao2024transformers}. However for the purposes of this work, we treat the Mamba2 block as a combination of an input linear projection ($W_{in}$, which can be broken down into $W_z$, $W_x$, $W_B$, $W_C$, $W_{dt}$), a causal 1D convolution layer with kernel size 4 (with weights that are a concatenation of $W_{conv_x}$, $W_{conv_B}$, and $W_{conv_C}$ applied in groups), a chunk + selective scan operation ($\textit{SSM}$), and an output projection layer ($W_{out}$). Similar to a Transformer, this block takes in an $(b, l, d)$ shaped tensor -- $b$ is batch size, $l$ is sequence length, and $d$ is the dimensionality -- and produces a $(b, l, d)$ shaped output after a sequence transformation. For an input tensor $u$, the Mamba2 block $M(u)$ consists of the following steps:
\begin{equation}
    \textit{XBC}(u) = \sigma(\textit{Conv}(W_{conv_x}{^\frown}W_{conv_B}{^\frown}W_{conv_C}, W_{x}.u{^\frown}W_{B}.u{^\frown}W_{C}.u))
\end{equation}
\begin{equation}
    Y(u) = \textit{SSM}(\textit{XBC}(u), W_{dt}.u, A, D)
\end{equation}
\begin{equation}
    M(u) = \textit{Norm}(Y(u).{\sigma}(W_{z}.u)).W_{out}^T
\end{equation}

where ${^\frown}$ is the concatenation operation, $\textit{Conv}(k, s)$ applies a 1-D causal convolution with weights $k$ (applied in $len(k)$ groups) on a sequence $s$, and $A$ and $D$ are learnable SSM parameters. $\sigma$ is a nonlinearity which we set to SiLU~\citep{elfwing2018sigmoid}, and $\textit{Norm}$ is a layer norm function which we set to RMSNorm~\citep{zhang2019root}.

\subsection{MatMamba Block}
A MatMamba block also has both input and output shapes as $(b,l,d)$. It is defined as a nested combination of $g$ Mamba2 blocks $M_i$, such that $M_1 \subset M_2 \subset ... \subset M_g$, where $M_i \subset M_j$ means that all the parameters of a sub-block $M_i$ are present in $M_j$. Works like MatFormer~\citep{kudugunta2023matformer}, OFA~\citep{cai2019once}, and Flextron~\citep{cai2024flextron} all share similar designs in which the largest model $M_g$ is the single universal base model from which numerous smaller submodels $M_i$ can be flexibly extracted. In MatMamba, we impose the nested structure along the \textit{dimensions} of the model parameters. Specifically for a sub-block $M_i$ with expansion factor $e = \frac{d_{inner}}{d_{model}}$, we choose a Matryoshka dimension $m_i$, such that $0 < m_i < d_{model}$, which results in an inner slice dimension $d_i = e \times m_i$ and number of heads $h_{i} = \frac{d_i}{d_{head}}$, subject to $d_i\mod d_{head} = 0$. For example, parameters like $W_x$ have a shape of $(d_{inner}, d_{model})$. For the $M_i$ sub-block, it will become $W_x[0:d_i]$ by slicing it along the $d_{inner}$ dimension. Similarly for parameters like A which have a shape of $(n_{heads})$, it will become $A[0:h_i]$. Concretely, the MatMamba block $M_{i}(u)$ when applied to an input tensor $u$ is these steps:

\begin{align}
    \textit{XBC}_{i}(u) = \sigma(\textit{Conv}(W_{conv_x}[0:d_i]{^\frown}W_{conv_B}{^\frown}W_{conv_C}, W_{x}[0:d_i].u{^\frown}W_{B}.u{^\frown}W_{C}.u))
\end{align}
\begin{equation}
    Y_{i}(u) = \textit{SSM}(\textit{XBC}_{i}(u), W_{dt}[0:h_i].u, A[0:h_i], D[0:h_i])
\end{equation}
\begin{equation}
    M_{i}(u) = \textit{Norm}(Y_{i}(u).{\sigma}(W_{z}[0:d_{i}].u)).W_{out}[0:d_i]^T
\end{equation}



In practice, $W_z$, $W_x$, $W_B$, $W_C$, and $W_{dt}$ are implemented as a single input projection layer with tensor parallelism, with appropriate rearranging of dimensions depending on $m_i$. Figure \ref{fig:matmamba-block} illustrates the MatMamba block. We also provide PyTorch-style pseudocode for the block in Appendix \ref{sec:A}, to provide a clearer understanding of our implementation. 

Compared to MatFormer~\citep{kudugunta2023matformer}, where the Matryoshka structure is only applied on the MLP subblock of the Transformer block, MatMamba applies nesting to the entire block wherever the inner dimension plays a role. This leads to a nearly linear reduction in total parameter count (and also a nearly linear reduction in flop count due to the nature of Mamba2). Also, typically $>95\%$ of the parameter count in a MatMamba block is in the input and output projections, which can be converted into nested layers while maintaining the well understood systems characteristics of projection layers. See Appendix \ref{sec:A} for a detailed example of parameter count reduction.

We can stack $L$ such MatMamba blocks to create a MatMamba model. For a given $m_i$ and nested blocks $M_1 \subset M_2 \subset ... \subset M_g$, we can create a MatMamba model $f_i$ with $L$ layers, and $g$ nested models $f_1 \subset f_2 \subset ... \subset f_g$. Each $f_i$ is formed by stacking $M_i$ $L$ times. Like Mamba2, the MatMamba backbone is a general purpose sequence processing architecture, which with an appropriate tokenizer and output head can process a variety of modalities.

\subsection{Training}
To train a model comprised of MatMamba blocks for $g$ chosen granularities, we perform $g$ forward passes to calculate a joint loss function. For an input $x$, model $f$, target $y$ and loss function $\mathcal{L}$:
\begin{equation}
    \mathcal{L}_{joint}(x, y) = \sum_{i=1}^{g}\lambda_{i} . \mathcal{L}(f_{i}(x), y)
\end{equation}
where $\lambda_i$ is the weight of the $i$-th nested submodel's loss. In this work, we train $g=4$ nested submodels with a uniform $\lambda_i = 1/g = 0.25$ for each submodel. As shown in Figure \ref{fig:matmamba-block}, during each forward pass, we accumulate gradients. The parameter update is done with a single backward pass. During the whole process, the model and the weights are the same, thereby also making memory usage the same as a regular Mamba2 block. In this work, we train MatMamba models with $g=4$ nested granularities, with the corresponding list of $m_i$'s being $[d_{model}, d_{model}/2, d_{model}/4, d_{model}/8]$, i.e. a halving of dimensionality for every sub-model.
Like MatFormer~\citep{kudugunta2023matformer} and Flextron~\citep{cai2024flextron}, we note that it is also possible to finetune an existing pretrained model to produce a nested structure. However, in this work, we focus on training from scratch to study the scaling characteristics of MatMamba models.
\subsection{Mix'n'Match\label{sec:mixnmatch}}
We can apply the Mix'n'Match strategy from MatFormer~\citep{kudugunta2023matformer} to flexibly extract \textit{any} submodel from MatMamba for inference. Concretely, for a model $f$ with $L$ layers, we need to choose a dimensionality $m_{i}$ at each layer $i$. Note that $m_{i}$ can be either one of the explicitly optimized $g$ granularites (e.g. picking from one of [1024, 512, 256, 128] from a 135M-MatMamba-Vision model, see section \ref{sec:matmambavision}), or we can choose interpolated dimensionalities that were not explicitly optimized for (e.g. picking any random valid value like 768 or 384 that was not explicitly trained). For instance, we could choose $m_{1}=256$ (25\% size) in layer 1, $m_{2}=1024$ (100\% size) in layer 2, $m_{3}=768$ (75\% size) in layer 3, and so on. The only constraint on $m_{i}$ in MatMamba is that it needs to lead to an integer number of heads, or that $(e\times{m_{i}})\mod d_{head} = 0$, where $e=\frac{d_{inner}}{d_{model}}$. This leads to a combinatorially large number of possible submodels (beyond the $g$ explicitly optimized granularities) that can be flexibly extracted -- all from the same set of base model weights --  as shown in Figure \ref{fig:matmamba-block}. Due to the Matryoshka structure, the first few dimensions (that are shared among all the nested submodels) are incentivized to learn the strongest representations. 

\subsection{Elastic Inference}
When deploying a MatMamba model for inference, we typically need to store the single universal model $f_g$ in memory. If compute is not constrained (or if the inference workload is predictable), then we can use the full model to get the most accurate results. However, depending on dynamic constraints (e.g. available inference compute, energy usage, system load, desired accuracy etc.), we can perform a forward pass on a chosen slice of the network on the fly.

There are exciting possibilities like combining cloud and edge inference -- we could store a smaller model $f_i$ on the edge device and when necessary, use the larger model $f_j$ on the cloud, or using a smaller model to act as a draft model for speculative decoding~\citep{leviathan2023fast} with a larger verifier model. We could also potentially do input-adaptive sub-model selection (e.g. use a larger model for a more difficult input). All of these are possible only because MatMamba has a consistent and nested Matryoshka structure, in which all the sub-models share the same metric space.

\section{Experiments}
In this section, we demonstrate the effectiveness of MatMamba-based models across two modalities: vision (\textbf{MatMamba-Vision}) and language (\textbf{MatMamba-LM}). For vision, we show results for image classification (Section \ref{sec:imageclassification}) and adaptive image retrieval (Section \ref{sec:imageretrieval}). For language, we train decoder language models (Section \ref{sec:lm}). We train models at a variety of scales from 35M to 1.4B parameters.
For a fair comparison, we also independently train baseline Mamba2 models which have the same architecture as the submodels of each MatMamba granularity. Please note that \textit{we do not aim to achieve state-of-the-art results} in this work on either language or vision for the chosen model sizes. We instead focus on properties like nested structure consistency, parameter reduction, inference speedups/memory usage for submodels, and scaling of simple networks built using the MatMamba block.

\subsection{MatMamba-Vision\label{sec:matmambavision}}

\begin{figure}[t!]
    \centering
    \includegraphics[width=\linewidth]{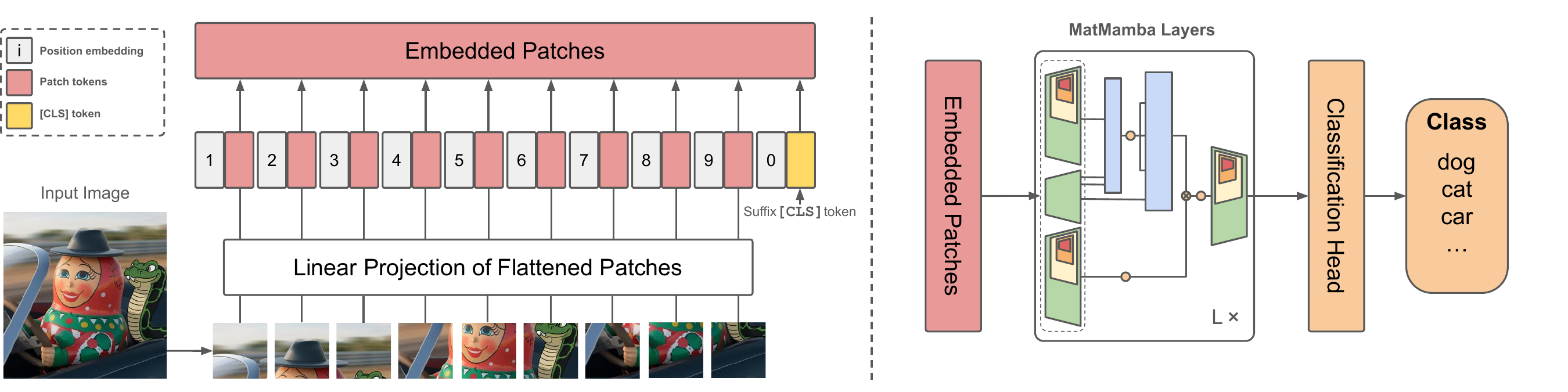}
    \caption{MatMamba layers for vision tasks. Similar to a ViT~\citep{dosovitskiy2020image}, we convert an image into a tensor of embedded patches. Because of the causal nature of the Mamba2 block, we suffix the [\texttt{CLS}] token. We intentionally keep the design simple to better study the properties of the MatMamba block.}
    \label{fig:matmamba-vision-overview}
\end{figure}

\begin{table}[b!]
    \centering
    \caption{Base model architectures for MatMamba-Vision (35M and 135M) with the explicitly optimized submodels for $g=4$ nested granularities.}
    \label{tab:base_arch_vision}
    \begin{tabular}{ccccrr}
        \toprule
         & & & & \multicolumn{2}{c}{Parameters}\\
        Base Model & Layers & $m_i$ & $h_i$ & Patch embed & MatMamba Layers \\ 
        \midrule
        \multirow{4}{*}{135M-1024D} & 20 & 1024 & 32 & 787,456 & 132,739,840 \\
         & 20 & 512 & 16 & 787,456 & 69,004,160 \\
         & 20 & 256 & 8 & 787,456 & 37,136,320 \\
         & 20 & 128 & 4 & 787,456 & 21,202,400 \\
        \midrule
        \multirow{4}{*}{35M-512D} & 20 & 512 & 32 & 393,728 &  34,927,360 \\
         & 20 & 256 & 16 & 393,728 &  18,787,200 \\
         & 20 & 128 & 8 & 393,728 &  10,717,120 \\
         & 20 & 64 & 4 & 393,728 &  6,682,080 \\
        \bottomrule
    \end{tabular}
\end{table}
MatMamba-Vision (Figure \ref{fig:matmamba-vision-overview}) contains a patch embedding followed by $L$ MatMamba blocks with a unidirectional SSM scan. One crucial design choice we make is to use the \texttt{[CLS]} token as a suffix instead of the conventional prefix. This allows it to attend to information from the entire sequence. We find that this simple architecture works effectively on both image classification and adaptive retrieval. We train two model variations (35M with $d_{model}=512$ and 135M with $d_{model}=1024$, see Table \ref{tab:base_arch_vision}) with patch size 16 and $L=20$ layers on ImageNet-1k \cite{deng2009imagenet} which has 1.28M training images and 50k validation images.
Compared to other recent work on SSM's for vision tasks like MambaVision~\citep{hatamizadeh2024mambavision}, MambaND~\citep{li2024mamba}, and Vision Mamba~\citep{zhu2024vision} -- all of which have major design changes on top of Mamba layers like bidirectional scan with additional projections, varying order of scans, or combining SSM layers with attention and convolution layers -- we keep our network architecture as simple as possible. 
We use FFCV~\citep{leclerc2023ffcv} dataloaders for efficient training. We apply augmentations like RandAug~\citep{cubuk2020randaugment}, Random Erasing~\citep{zhong2020random}, Mixup~\citep{zhang2017mixup}, Cutmix~\citep{yun2019cutmix}, and a number of other settings following DEiT-3~\citep{touvron2022deit}, AugReg~\citep{steiner2021train}, and Better ViT Baselines~\citep{beyer2022better}.
The exact detailed experimental settings can be seen in Appendix \ref{sec:A}.

\begin{figure}[t!]
    \centering
    \begin{subfigure}[b]{0.49\textwidth}
        \includegraphics[width=\textwidth]{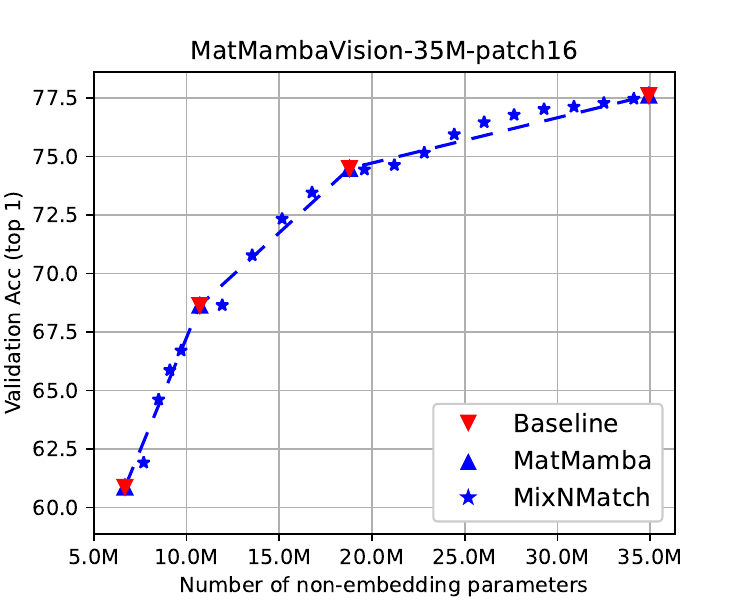}
        \caption{35M model}
        \label{fig:imagenetsubfig1}
    \end{subfigure}
    \hfill
    \begin{subfigure}[b]{0.49\textwidth}
        \includegraphics[width=\textwidth]{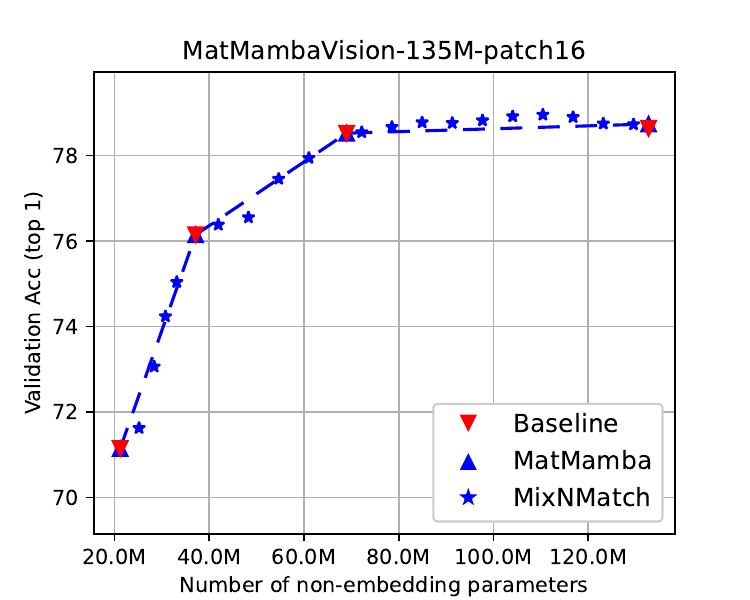}
        \caption{135M model}
        \label{fig:imagenetsubfig2}
    \end{subfigure}
    \caption{ImageNet-1K Classification: MatMamba-Vision is as accurate as explicitly trained baselines across various constraints while also spanning the accuracy-vs-compute pareto optimal curve through mix'n'match submodels.}
    \label{fig:imagenet-classification}
\end{figure}

\subsubsection{Image Classification\label{sec:imageclassification}}
In Figure \ref{fig:imagenet-classification}, we see that for both the 35M and 135M MatMamba-Vision models, the explicitly optimized submodels closely match the 4 independently trained baseline models with the same architecture as the nested submodel. However, instead of needing four separate models, we can get all levels of performance/parameter counts flexibly in a single model.

\begin{figure}[b!]
    \centering
    \begin{subfigure}[b]{0.49\textwidth}
        \includegraphics[width=\textwidth]{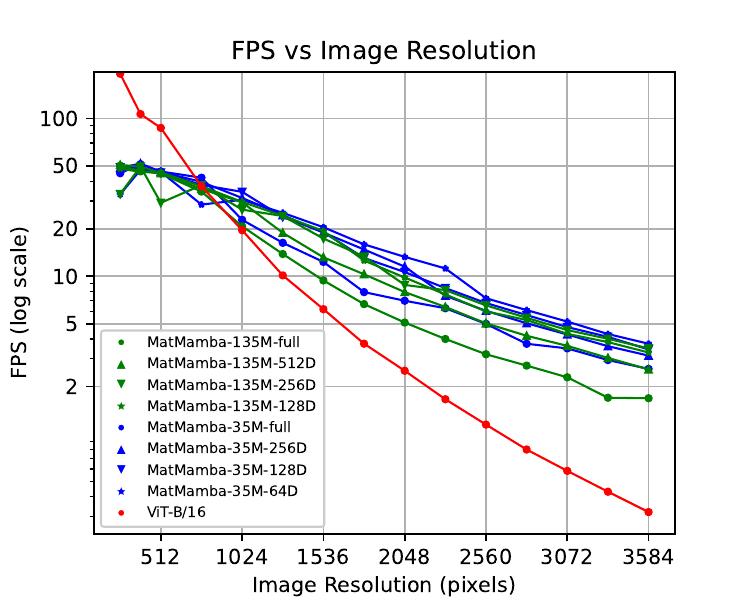}
        \caption{MatMamba-Vision FPS vs. Image Size}
        \label{fig:imagenetsubfig5}
    \end{subfigure}
    \hfill
    \begin{subfigure}[b]{0.49\textwidth}
        \includegraphics[width=\textwidth]{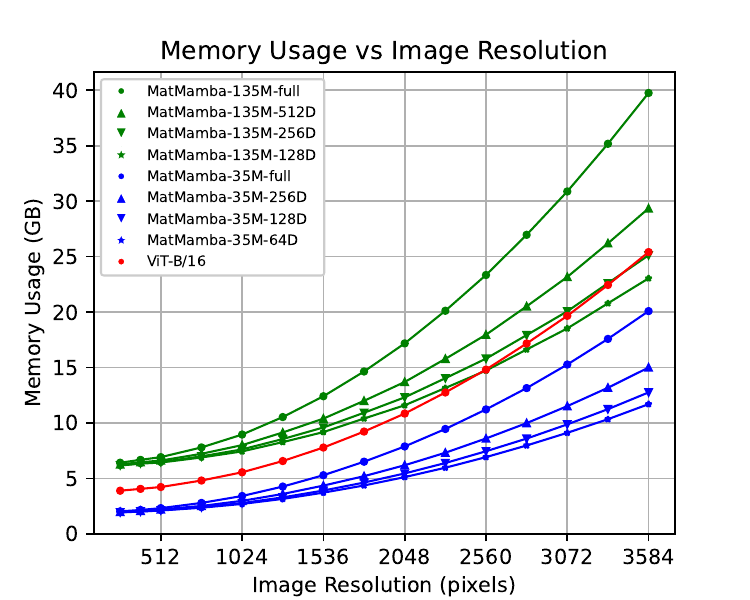}
        \caption{MatMamba-Vision Memory Usage}
        \label{fig:imagenetsubfig6}
    \end{subfigure}
    \caption{Inference speed and memory usage for batch size 1 on an H100 for nested MatMamba-Vision models and a ViT baseline. At larger resolutions, the characteristics of MatMamba are better.}
    \label{fig:imagenet-profiling}
\end{figure}
\textbf{Adaptive Inference using Mix'n'Match:} Additionally (Figure \ref{fig:imagenet-classification}), using Mix'n'Match at a variety of combined granularities yields models that smoothly interpolate (and sometimes exceed) the accuracy on the line joining the explicitly optimized granularities. This points towards powerful adaptivity, because we can extract a combinatorially large number of submodels along the accuracy-compute curve. We can optimize submodel selection for deployment constraints flexibly, all while only using the weights of a single nested universal model.

\textbf{Inference Speeds at Higher Resolutions:} In Figure \ref{fig:imagenet-profiling}, we also study the inference speed tradeoffs of nested granularities of MatMamba-Vision models when compared with each other and a ViT-B/16 model. We find that at or below 512px resolution, the sequence length is low enough for the ViT to be the fastest model (due to GPU parallelism and optimizations like FlashAttention). However, as we increase the resolution to 1024px and beyond, Mamba-style models start outperforming ViT in both througput and latency. We also study inference memory usage, and find that MatMamba-Vision scales slightly better than an optimized ViT-B/16 as the resolution increases. Both of these observations offer promising evidence that MatMamba-based models can be suitable for processing longer sequences of visual data at higher resolutions on a single accelerator (as opposed to scaling context length in Transformers using methods like RingAttention \cite{liu2023ring} which needs multiple interconnected accelerators for a single forward pass at long sequence lengths).

\begin{figure}[t!]
    \centering
    \begin{subfigure}[b]{0.49\textwidth}
        \includegraphics[width=\textwidth]{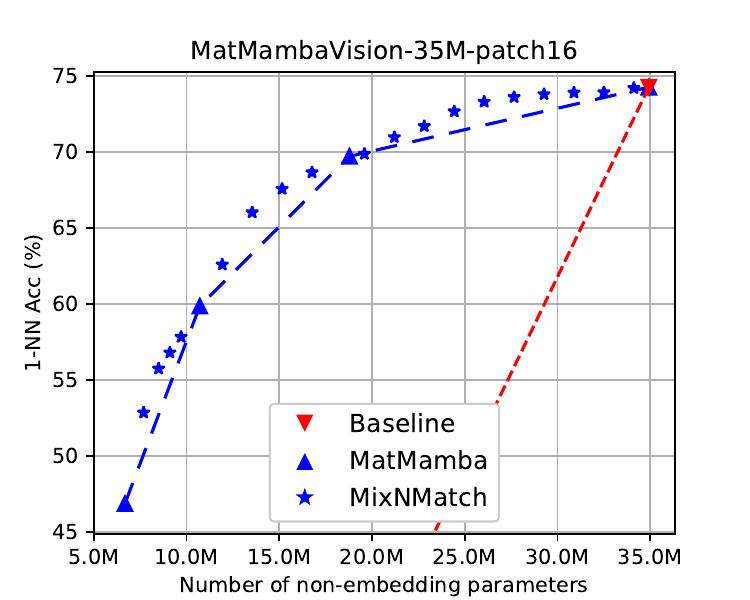}
        \caption{1-NN Retrieval for 35M model}
        \label{fig:imagenetsubfig3}
    \end{subfigure}
    \hfill
    \begin{subfigure}[b]{0.49\textwidth}
        \includegraphics[width=\textwidth]{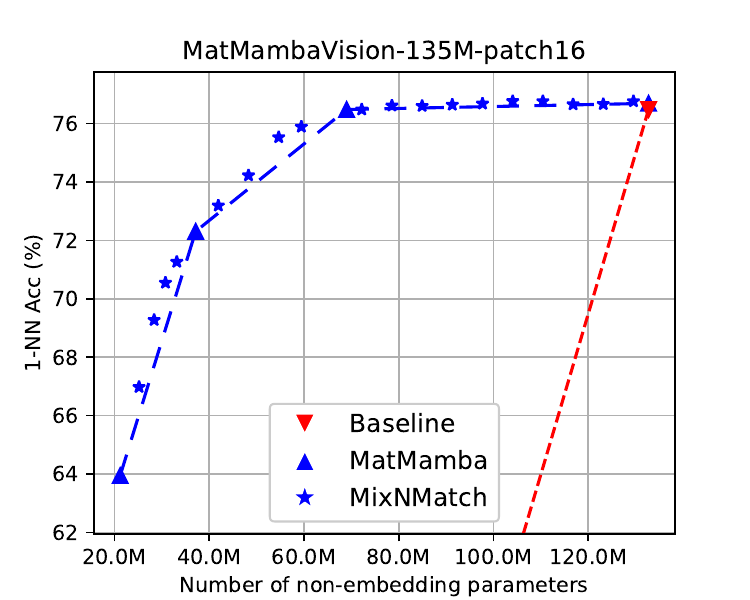}
        \caption{1-NN Retrieval for 135M model}
        \label{fig:imagenetsubfig4}
    \end{subfigure}
    \caption{Adaptive Image Retrieval on ImageNet-1K: Submodels obtained from the largest MatMamba-Vision model preserve the metric space of embeddings resulting in accurate and adaptive query processing at scale while baseline struggles to work across models without distillation.}
    \label{fig:imagenet-retrieval}
\end{figure}
\subsubsection{Adaptive Image Retrieval\label{sec:imageretrieval}}
Image retrieval aims to locate semantically similar images using representations generated by a pretrained encoder~\citep{chen2022deep}. The standard method involves encoding both database and query images with the same encoder and then performing nearest neighbor retrieval. While using a powerful encoder for database images is feasible, the query encoder must be efficient for real-time applications. Moreover, query encoding scenarios can vary, such as on-device versus cloud processing and varying query load and complexity. Existing solutions with fixed encoders often compromise accuracy or cost in different settings.

Due to its flexibility, MatMamba-Vision is a promising candidate for query encoding. However, retrieval also requires that submodels maintain distance relationships between fixed database (encoded with a larger encoder) and query embeddings across various granularities. Using smaller baseline Mamba2 models solely for query encoding can lead to significant distance preservation issues and poor retrieval accuracy (as illustrated in Figure~\ref{fig:imagenet-retrieval}).

We evaluated both the baseline and MatMamba-Vision encoders on ImageNet-1K for image retrieval at 35M and 135M parameter scales. Using the [\texttt{CLS}] token representation, we calculated 1-nearest neighbor (NN) accuracy. Figure~\ref{fig:imagenet-retrieval} demonstrates that submodels extracted from MatMamba can effectively preserve distances and offer greater flexibility. For example, MatMamba-Vision-135M can reduce compute cost by 55\% with a minimal accuracy loss of less than 0.5\%. While causal models with suffix [\texttt{CLS}] token might not be as accurate as bi-directional encoders for retrieval, this is a promising start towards better long-context encoders while enabling adaptive query processing.

\subsection{MatMamba-LM\label{sec:lm}}
\begin{table}[t!]
    \centering
    \caption{Base model architectures for MatMamba-LM}
    \label{tab:base_arch}
    \begin{tabular}{ccccrrr}
        \toprule
        Base Model & Layers & $d_{model}$ & $d_{head}$ & Embed params & Non-embed params & Tokens\\ 
        \midrule
        130M & 24 & 768 & 24 & 38,615,040 & 90,368,448 & 62.9B\\
        370M & 48 & 1024 & 32 & 51,486,720 & 316,851,712 & 125.8B\\
        790M & 48 & 1536 & 48 & 77,230,080 & 702,918,912 & 125.8B\\
        1.4B & 48 & 2048 & 64 & 102,973,440 & 1,240,767,488 & 251.6B\\
        \bottomrule
    \end{tabular}
\end{table}

We train decoder language models using the MatMamba block (MatMamba-LM). The models closely follow the training procedure and hyperparameters of \texttt{llm.c}~\citep{karpathy2024llm}. We use the GPT-2~\citep{radford2019language} tokenizer with a padded vocabulary size of 50,280. We use the FineWeb~\citep{penedo2024fineweb} dataset to train all models. We train 4 separate models (with base model parameter sizes 130M, 370M, 790M, and 1.4B). For each of these base models, we optimize $g=4$ nested granularities $[d_{model}, d_{model}/2, d_{model}/4, d_{model}/8]$. For baselines, we train vanilla Mamba2 models with the same architecture as the nested submodels. Table \ref{tab:base_arch} shows the exact configurations for each model.

\begin{figure}[b!]
    \centering
    \begin{subfigure}[b]{0.49\textwidth}
        \includegraphics[width=\textwidth]{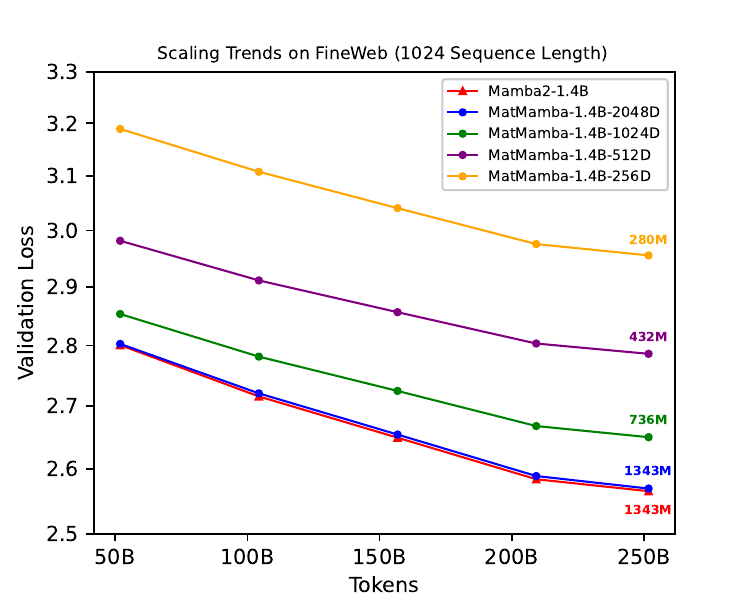}
        \caption{1.4B model}
        \label{fig:lmscaling1}
    \end{subfigure}
    \hfill
    \begin{subfigure}[b]{0.49\textwidth}
        \includegraphics[width=\textwidth]{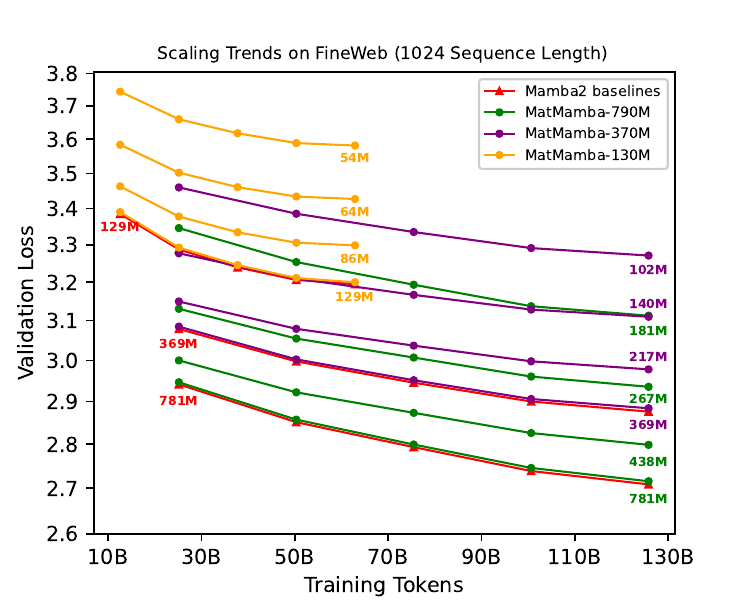}
        \caption{All other models}
        \label{fig:lmscaling2}
    \end{subfigure}
    \caption{MatMamba-LM scales as well as explictly optimized Mamba2 baselines across all model and training scales all while providing accurate sub-models on the go.}
    \label{fig:lm-scaling-curves}
\end{figure}
\begin{figure}
    \centering
    \begin{subfigure}[b]{0.49\textwidth}
        \includegraphics[width=\textwidth]{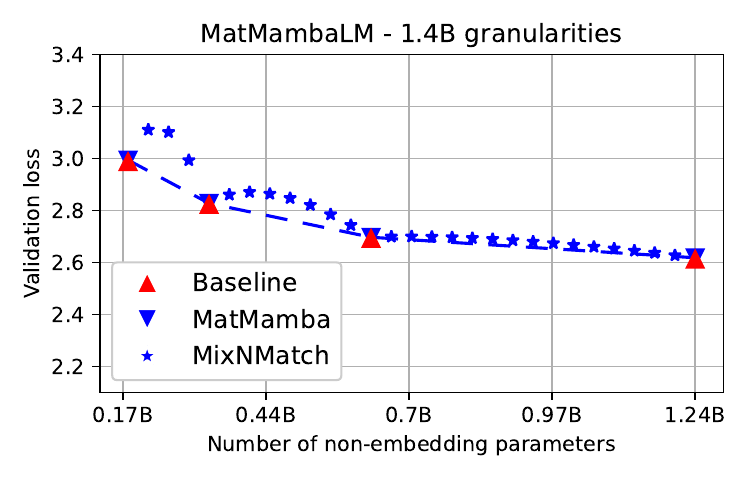}
        \caption{1.4B model}
        \label{fig:lmloss1}
    \end{subfigure}
    \hfill
    \begin{subfigure}[b]{0.49\textwidth}
        \includegraphics[width=\textwidth]{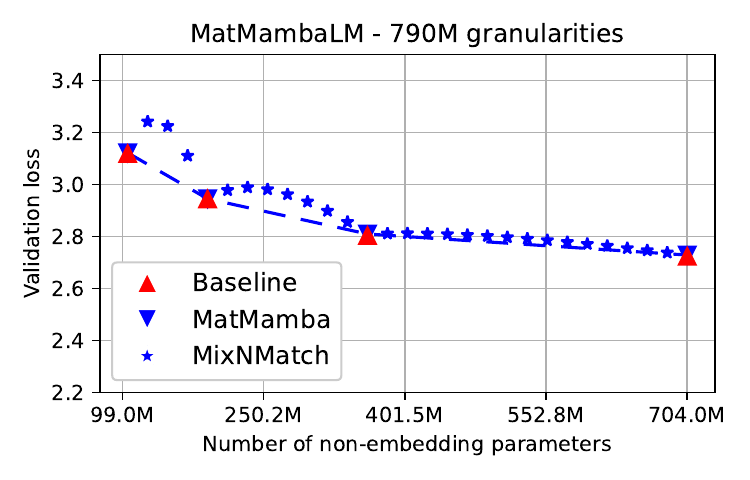}
        \caption{790M model}
        \label{fig:lmloss2}
    \end{subfigure}
    
    \vspace{0.5cm} 
    
    \begin{subfigure}[b]{0.49\textwidth}
        \includegraphics[width=\textwidth]{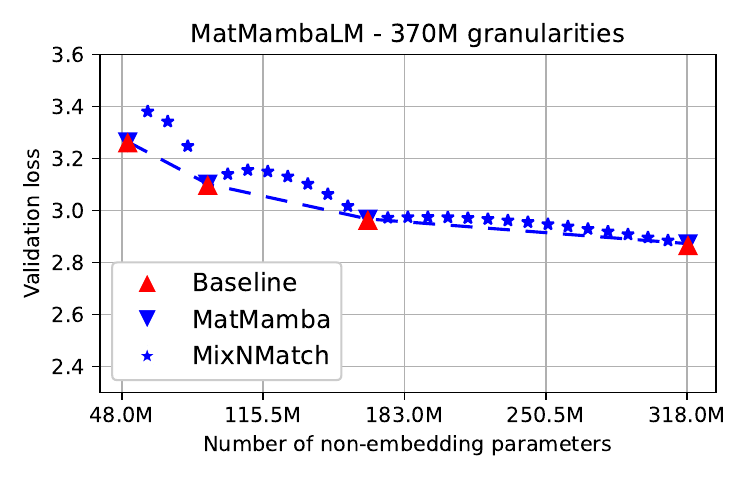}
        \caption{370M model}
        \label{fig:lmloss3}
    \end{subfigure}
    \hfill
    \begin{subfigure}[b]{0.49\textwidth}
        \includegraphics[width=\textwidth]{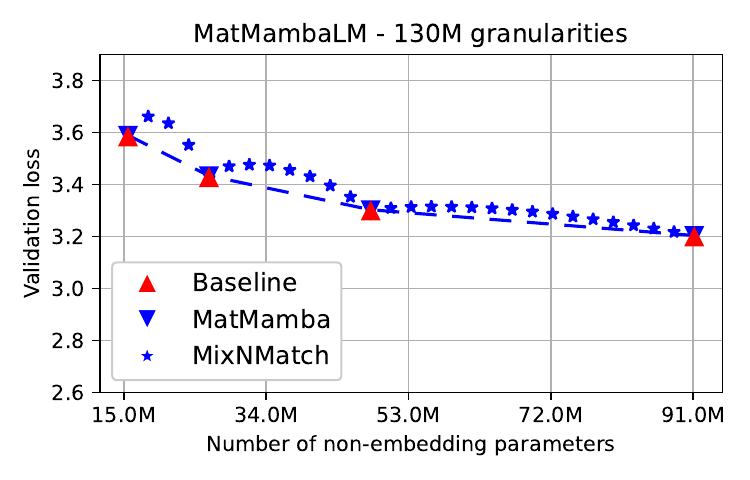}
        \caption{130M model}
        \label{fig:lmloss4}
    \end{subfigure}
    
    \caption{Validation loss for the language modelling task across model sizes showing MatMamba-LM is accurate as a Mamba2 baseline at explictely optimized granualrities, while enabling pareto optimal submodels through Mix'n'Match.}
    \label{fig:lm-loss}
\end{figure}

\textbf{MatMamba-LM scales as well as Mamba2:}
In Figure \ref{fig:lm-scaling-curves}, we see that MatMamba-LM models scale with training tokens as well as Mamba2 models for the largest granularity. In Figure \ref{fig:lm-loss}, we also see that for all granularities, the final trained models of every granularity scale as well as the baseline model trained with the same architecture. Furthermore, we observe that the validation loss in Figure \ref{fig:lm-scaling-curves} for every nested granularity is at a similar distance (usually a delta of 0.4 in val loss) between the largest model ($m_i=d_{model}$) and the smallest model ($m_i=d_{model}/8$), with consistent gaps for the intermediate models. Given that validation loss on a large and diverse dataset is the strongest proxy for language model performance (as opposed to few-shot evals on noisy datasets), these scaling trends offer very promising evidence that a single nested MatMamba-LM model can be used in a variety of deployments instead of training 4 separate models independently.

In Figure \ref{fig:lm-loss}, we show results for adaptive inference using Mix'n'Match on all 4 MatMamba-LM variants. We see a smooth interpolation between the $d_{model}/2$ and $d_{model}$ granularities (e.g. between $d_{model}/8$ and $d_{model}/4$). However, for the lower granularities, even though the explicitly optimized granularities scale as well as expected, the Mix'n'Match models that have not been explicitly trained suffer a slight performance degradation. We observed that during earlier stages of training, the Mix'n'Match trends for all granularities were exactly on the performance-compute curve. However, towards the later stages, the explicitly optimized granularities improve faster than the Mix'n'Match granularities (almost like anchor points). There are mechanisms that can potentially fix this: like a self-distillation loss with the output of the largest submodel, training with more than $g=4$ granularities, or the surrogate model structure used in Flextron~\citep{cai2024flextron}, that should make the Mix'n'Match trend smooth. However, this requires more rigorous understanding, and we leave deeper exploration to future work.

\section{Conclusions}
In this work, we presented MatMamba, which is a way to impose a nested Matryoshka strcuture on a Mamba2 state space model. It brings together the best of both Mamba-style models (faster inference times, especially for longer sequences) and Matryoshka-style learning. A single MatMamba model contains hundreds of nested and accurate submodels that can be flexibly extracted for inference. MatMamba-Vision and MatMamba-LM models match the performance and accuracy of the independently trained Mamba2 baselines. MatMamba models allow us to choose a desired performance-compute tradeoff, all while being a single Matryoshka-style model instead of multiple different models for specific scenarios. This enables interesting use cases like speculative decoding using a smaller draft model and a larger verifier model, input-adaptive submodel selection, and hybrid cloud-edge inference with the same model based on available compute.


\newpage

\bibliography{iclr2025_conference}

\begin{thebibliography}{53}
\providecommand{\natexlab}[1]{#1}
\providecommand{\url}[1]{\texttt{#1}}
\expandafter\ifx\csname urlstyle\endcsname\relax
  \providecommand{\doi}[1]{doi: #1}\else
  \providecommand{\doi}{doi: \begingroup \urlstyle{rm}\Url}\fi

\bibitem[Beck et~al.(2024)Beck, P{\"o}ppel, Spanring, Auer, Prudnikova, Kopp, Klambauer, Brandstetter, and Hochreiter]{beck2024xlstm}
Maximilian Beck, Korbinian P{\"o}ppel, Markus Spanring, Andreas Auer, Oleksandra Prudnikova, Michael Kopp, G{\"u}nter Klambauer, Johannes Brandstetter, and Sepp Hochreiter.
\newblock xlstm: Extended long short-term memory.
\newblock \emph{arXiv preprint arXiv:2405.04517}, 2024.

\bibitem[Beyer et~al.(2022)Beyer, Zhai, and Kolesnikov]{beyer2022better}
Lucas Beyer, Xiaohua Zhai, and Alexander Kolesnikov.
\newblock Better plain vit baselines for imagenet-1k.
\newblock \emph{arXiv preprint arXiv:2205.01580}, 2022.

\bibitem[Beyer et~al.(2023)Beyer, Izmailov, Kolesnikov, Caron, Kornblith, Zhai, Minderer, Tschannen, Alabdulmohsin, and Pavetic]{beyer2023flexivit}
Lucas Beyer, Pavel Izmailov, Alexander Kolesnikov, Mathilde Caron, Simon Kornblith, Xiaohua Zhai, Matthias Minderer, Michael Tschannen, Ibrahim Alabdulmohsin, and Filip Pavetic.
\newblock Flexivit: One model for all patch sizes.
\newblock In \emph{Proceedings of the IEEE/CVF Conference on Computer Vision and Pattern Recognition}, pp.\  14496--14506, 2023.

\bibitem[Botev et~al.(2024)Botev, De, Smith, Fernando, Muraru, Haroun, Berrada, Pascanu, Sessa, Dadashi, et~al.]{botev2024recurrentgemma}
Aleksandar Botev, Soham De, Samuel~L Smith, Anushan Fernando, George-Cristian Muraru, Ruba Haroun, Leonard Berrada, Razvan Pascanu, Pier~Giuseppe Sessa, Robert Dadashi, et~al.
\newblock Recurrentgemma: Moving past transformers for efficient open language models.
\newblock \emph{arXiv preprint arXiv:2404.07839}, 2024.

\bibitem[Cai et~al.(2019)Cai, Gan, Wang, Zhang, and Han]{cai2019once}
Han Cai, Chuang Gan, Tianzhe Wang, Zhekai Zhang, and Song Han.
\newblock Once-for-all: Train one network and specialize it for efficient deployment.
\newblock \emph{arXiv preprint arXiv:1908.09791}, 2019.

\bibitem[Cai et~al.(2024{\natexlab{a}})Cai, Yang, Gao, and Lee]{cai2024matryoshka}
Mu~Cai, Jianwei Yang, Jianfeng Gao, and Yong~Jae Lee.
\newblock Matryoshka multimodal models.
\newblock \emph{arXiv preprint arXiv:2405.17430}, 2024{\natexlab{a}}.

\bibitem[Cai et~al.(2024{\natexlab{b}})Cai, Muralidharan, Heinrich, Yin, Wang, Kautz, and Molchanov]{cai2024flextron}
Ruisi Cai, Saurav Muralidharan, Greg Heinrich, Hongxu Yin, Zhangyang Wang, Jan Kautz, and Pavlo Molchanov.
\newblock Flextron: Many-in-one flexible large language model.
\newblock \emph{arXiv preprint arXiv:2406.10260}, 2024{\natexlab{b}}.

\bibitem[CartesiaAI(2024)]{Sonic}
CartesiaAI.
\newblock {Sonic}, 2024.
\newblock URL \url{https://cartesia.ai/blog/sonic}.
\newblock [Online; accessed 10/01/2024].

\bibitem[Chen et~al.(2022)Chen, Liu, Wang, Bakker, Georgiou, Fieguth, Liu, and Lew]{chen2022deep}
Wei Chen, Yu~Liu, Weiping Wang, Erwin~M Bakker, Theodoros Georgiou, Paul Fieguth, Li~Liu, and Michael~S Lew.
\newblock Deep learning for instance retrieval: A survey.
\newblock \emph{IEEE Transactions on Pattern Analysis and Machine Intelligence}, 2022.

\bibitem[Cubuk et~al.(2020)Cubuk, Zoph, Shlens, and Le]{cubuk2020randaugment}
Ekin~D Cubuk, Barret Zoph, Jonathon Shlens, and Quoc~V Le.
\newblock Randaugment: Practical automated data augmentation with a reduced search space.
\newblock In \emph{Proceedings of the IEEE/CVF conference on computer vision and pattern recognition workshops}, pp.\  702--703, 2020.

\bibitem[Dao \& Gu(2024)Dao and Gu]{dao2024transformers}
Tri Dao and Albert Gu.
\newblock Transformers are ssms: Generalized models and efficient algorithms through structured state space duality.
\newblock \emph{arXiv preprint arXiv:2405.21060}, 2024.

\bibitem[De et~al.(2024)De, Smith, Fernando, Botev, Cristian-Muraru, Gu, Haroun, Berrada, Chen, Srinivasan, et~al.]{de2024griffin}
Soham De, Samuel~L Smith, Anushan Fernando, Aleksandar Botev, George Cristian-Muraru, Albert Gu, Ruba Haroun, Leonard Berrada, Yutian Chen, Srivatsan Srinivasan, et~al.
\newblock Griffin: Mixing gated linear recurrences with local attention for efficient language models.
\newblock \emph{arXiv preprint arXiv:2402.19427}, 2024.

\bibitem[Deng et~al.(2009)Deng, Dong, Socher, Li, Li, and Fei-Fei]{deng2009imagenet}
Jia Deng, Wei Dong, Richard Socher, Li-Jia Li, Kai Li, and Li~Fei-Fei.
\newblock Imagenet: A large-scale hierarchical image database.
\newblock In \emph{2009 IEEE conference on computer vision and pattern recognition}, pp.\  248--255. Ieee, 2009.

\bibitem[Devvrit et~al.(2023)Devvrit, Kudugunta, Kusupati, Dettmers, Chen, Dhillon, Tsvetkov, Hajishirzi, Kakade, Farhadi, Jain, et~al.]{kudugunta2023matformer}
F~Devvrit, Sneha Kudugunta, Aditya Kusupati, Tim Dettmers, Kaifeng Chen, Inderjit Dhillon, Yulia Tsvetkov, Hannaneh Hajishirzi, Sham Kakade, Ali Farhadi, Prateek Jain, et~al.
\newblock Matformer: Nested transformer for elastic inference.
\newblock \emph{arXiv preprint arXiv:2310.07707}, 2023.

\bibitem[Dosovitskiy(2020)]{dosovitskiy2020image}
Alexey Dosovitskiy.
\newblock An image is worth 16x16 words: Transformers for image recognition at scale.
\newblock \emph{arXiv preprint arXiv:2010.11929}, 2020.

\bibitem[Dubey et~al.(2024)Dubey, Jauhri, Pandey, Kadian, Al-Dahle, Letman, Mathur, Schelten, Yang, Fan, et~al.]{dubey2024llama}
Abhimanyu Dubey, Abhinav Jauhri, Abhinav Pandey, Abhishek Kadian, Ahmad Al-Dahle, Aiesha Letman, Akhil Mathur, Alan Schelten, Amy Yang, Angela Fan, et~al.
\newblock The llama 3 herd of models.
\newblock \emph{arXiv preprint arXiv:2407.21783}, 2024.

\bibitem[Elfwing et~al.(2018)Elfwing, Uchibe, and Doya]{elfwing2018sigmoid}
Stefan Elfwing, Eiji Uchibe, and Kenji Doya.
\newblock Sigmoid-weighted linear units for neural network function approximation in reinforcement learning.
\newblock \emph{Neural networks}, 107:\penalty0 3--11, 2018.

\bibitem[Gu \& Dao(2023)Gu and Dao]{gu2023mamba}
Albert Gu and Tri Dao.
\newblock Mamba: Linear-time sequence modeling with selective state spaces.
\newblock \emph{arXiv preprint arXiv:2312.00752}, 2023.

\bibitem[Gu et~al.(2023)Gu, Zhai, Zhang, Susskind, and Jaitly]{gu2023matryoshka}
Jiatao Gu, Shuangfei Zhai, Yizhe Zhang, Joshua~M Susskind, and Navdeep Jaitly.
\newblock Matryoshka diffusion models.
\newblock In \emph{The Twelfth International Conference on Learning Representations}, 2023.

\bibitem[Hatamizadeh \& Kautz(2024)Hatamizadeh and Kautz]{hatamizadeh2024mambavision}
Ali Hatamizadeh and Jan Kautz.
\newblock Mambavision: A hybrid mamba-transformer vision backbone.
\newblock \emph{arXiv preprint arXiv:2407.08083}, 2024.

\bibitem[Hu et~al.(2024)Hu, Dou, Li, Kamath, Peng, and Chang]{hu2024matryoshka}
Wenbo Hu, Zi-Yi Dou, Liunian~Harold Li, Amita Kamath, Nanyun Peng, and Kai-Wei Chang.
\newblock Matryoshka query transformer for large vision-language models.
\newblock \emph{arXiv preprint arXiv:2405.19315}, 2024.

\bibitem[Jain et~al.(2024)Jain, Hegde, Kusupati, Nagrani, Buch, Jain, Arnab, and Paul]{jain2024mixture}
Gagan Jain, Nidhi Hegde, Aditya Kusupati, Arsha Nagrani, Shyamal Buch, Prateek Jain, Anurag Arnab, and Sujoy Paul.
\newblock Mixture of nested experts: Adaptive processing of visual tokens.
\newblock \emph{arXiv preprint arXiv:2407.19985}, 2024.

\bibitem[Jaiswal et~al.(2023)Jaiswal, Gan, Du, Zhang, Wang, and Yang]{jaiswal2023compressing}
Ajay Jaiswal, Zhe Gan, Xianzhi Du, Bowen Zhang, Zhangyang Wang, and Yinfei Yang.
\newblock Compressing llms: The truth is rarely pure and never simple.
\newblock \emph{arXiv preprint arXiv:2310.01382}, 2023.

\bibitem[Karpathy(2024)]{karpathy2024llm}
Andrej Karpathy.
\newblock llm.c: Llm training in simple, raw c/cuda.
\newblock \url{https://github.com/karpathy/llm.c}, 2024.
\newblock Accessed: 10/1/2024.

\bibitem[Katharopoulos et~al.(2020)Katharopoulos, Vyas, Pappas, and Fleuret]{katharopoulos2020transformers}
Angelos Katharopoulos, Apoorv Vyas, Nikolaos Pappas, and Fran{\c{c}}ois Fleuret.
\newblock Transformers are rnns: Fast autoregressive transformers with linear attention.
\newblock In \emph{International conference on machine learning}, pp.\  5156--5165. PMLR, 2020.

\bibitem[Kusupati(2024)]{kusupati2024towards}
Aditya Kusupati.
\newblock \emph{Towards Adaptive Intelligence}.
\newblock PhD thesis, University of Washington, 2024.

\bibitem[Kusupati et~al.(2022)Kusupati, Bhatt, Rege, Wallingford, Sinha, Ramanujan, Howard-Snyder, Chen, Kakade, Jain, et~al.]{kusupati2022matryoshka}
Aditya Kusupati, Gantavya Bhatt, Aniket Rege, Matthew Wallingford, Aditya Sinha, Vivek Ramanujan, William Howard-Snyder, Kaifeng Chen, Sham Kakade, Prateek Jain, et~al.
\newblock Matryoshka representation learning.
\newblock \emph{Advances in Neural Information Processing Systems}, 35:\penalty0 30233--30249, 2022.

\bibitem[Leclerc et~al.(2023)Leclerc, Ilyas, Engstrom, Park, Salman, and Madry]{leclerc2023ffcv}
Guillaume Leclerc, Andrew Ilyas, Logan Engstrom, Sung~Min Park, Hadi Salman, and Aleksander Madry.
\newblock Ffcv: Accelerating training by removing data bottlenecks.
\newblock In \emph{Proceedings of the IEEE/CVF Conference on Computer Vision and Pattern Recognition}, pp.\  12011--12020, 2023.

\bibitem[Leviathan et~al.(2023)Leviathan, Kalman, and Matias]{leviathan2023fast}
Yaniv Leviathan, Matan Kalman, and Yossi Matias.
\newblock Fast inference from transformers via speculative decoding.
\newblock In \emph{International Conference on Machine Learning}, pp.\  19274--19286. PMLR, 2023.

\bibitem[Li et~al.(2024{\natexlab{a}})Li, Li, Wang, He, Wang, Wang, and Qiao]{li2024videomamba}
Kunchang Li, Xinhao Li, Yi~Wang, Yinan He, Yali Wang, Limin Wang, and Yu~Qiao.
\newblock Videomamba: State space model for efficient video understanding.
\newblock \emph{arXiv preprint arXiv:2403.06977}, 2024{\natexlab{a}}.

\bibitem[Li et~al.(2024{\natexlab{b}})Li, Singh, and Grover]{li2024mamba}
Shufan Li, Harkanwar Singh, and Aditya Grover.
\newblock Mamba-nd: Selective state space modeling for multi-dimensional data.
\newblock \emph{arXiv preprint arXiv:2402.05892}, 2024{\natexlab{b}}.

\bibitem[Lieber et~al.(2024)Lieber, Lenz, Bata, Cohen, Osin, Dalmedigos, Safahi, Meirom, Belinkov, Shalev-Shwartz, et~al.]{lieber2024jamba}
Opher Lieber, Barak Lenz, Hofit Bata, Gal Cohen, Jhonathan Osin, Itay Dalmedigos, Erez Safahi, Shaked Meirom, Yonatan Belinkov, Shai Shalev-Shwartz, et~al.
\newblock Jamba: A hybrid transformer-mamba language model.
\newblock \emph{arXiv preprint arXiv:2403.19887}, 2024.

\bibitem[Liu et~al.(2023)Liu, Zaharia, and Abbeel]{liu2023ring}
Hao Liu, Matei Zaharia, and Pieter Abbeel.
\newblock Ring attention with blockwise transformers for near-infinite context.
\newblock \emph{arXiv preprint arXiv:2310.01889}, 2023.

\bibitem[Liu et~al.(2024)Liu, Zhang, and Zhang]{liu2024vision}
Xiao Liu, Chenxu Zhang, and Lei Zhang.
\newblock Vision mamba: A comprehensive survey and taxonomy.
\newblock \emph{arXiv preprint arXiv:2405.04404}, 2024.

\bibitem[Penedo et~al.(2024)Penedo, Kydl{\'\i}{\v{c}}ek, Lozhkov, Mitchell, Raffel, Von~Werra, Wolf, et~al.]{penedo2024fineweb}
Guilherme Penedo, Hynek Kydl{\'\i}{\v{c}}ek, Anton Lozhkov, Margaret Mitchell, Colin Raffel, Leandro Von~Werra, Thomas Wolf, et~al.
\newblock The fineweb datasets: Decanting the web for the finest text data at scale.
\newblock \emph{arXiv preprint arXiv:2406.17557}, 2024.

\bibitem[Peng et~al.(2023)Peng, Alcaide, Anthony, Albalak, Arcadinho, Biderman, Cao, Cheng, Chung, Grella, et~al.]{peng2023rwkv}
Bo~Peng, Eric Alcaide, Quentin Anthony, Alon Albalak, Samuel Arcadinho, Stella Biderman, Huanqi Cao, Xin Cheng, Michael Chung, Matteo Grella, et~al.
\newblock Rwkv: Reinventing rnns for the transformer era.
\newblock \emph{arXiv preprint arXiv:2305.13048}, 2023.

\bibitem[Qin et~al.(2024)Qin, Yang, Sun, Shen, Li, Sun, and Zhong]{qin2024hgrn2}
Zhen Qin, Songlin Yang, Weixuan Sun, Xuyang Shen, Dong Li, Weigao Sun, and Yiran Zhong.
\newblock Hgrn2: Gated linear rnns with state expansion.
\newblock \emph{arXiv preprint arXiv:2404.07904}, 2024.

\bibitem[Radford et~al.(2019)Radford, Wu, Child, Luan, Amodei, Sutskever, et~al.]{radford2019language}
Alec Radford, Jeffrey Wu, Rewon Child, David Luan, Dario Amodei, Ilya Sutskever, et~al.
\newblock Language models are unsupervised multitask learners.
\newblock \emph{OpenAI blog}, 1\penalty0 (8):\penalty0 9, 2019.

\bibitem[Rippel et~al.(2014)Rippel, Gelbart, and Adams]{rippel2014learning}
Oren Rippel, Michael Gelbart, and Ryan Adams.
\newblock Learning ordered representations with nested dropout.
\newblock In \emph{International Conference on Machine Learning}, pp.\  1746--1754. PMLR, 2014.

\bibitem[Steiner et~al.(2021)Steiner, Kolesnikov, Zhai, Wightman, Uszkoreit, and Beyer]{steiner2021train}
Andreas Steiner, Alexander Kolesnikov, Xiaohua Zhai, Ross Wightman, Jakob Uszkoreit, and Lucas Beyer.
\newblock How to train your vit? data, augmentation, and regularization in vision transformers.
\newblock \emph{arXiv preprint arXiv:2106.10270}, 2021.

\bibitem[Sun et~al.(2024)Sun, Li, Dalal, Xu, Vikram, Zhang, Dubois, Chen, Wang, Koyejo, et~al.]{sun2024learning}
Yu~Sun, Xinhao Li, Karan Dalal, Jiarui Xu, Arjun Vikram, Genghan Zhang, Yann Dubois, Xinlei Chen, Xiaolong Wang, Sanmi Koyejo, et~al.
\newblock Learning to (learn at test time): Rnns with expressive hidden states.
\newblock \emph{arXiv preprint arXiv:2407.04620}, 2024.

\bibitem[Sun et~al.(2023)Sun, Dong, Huang, Ma, Xia, Xue, Wang, and Wei]{sun2023retentive}
Yutao Sun, Li~Dong, Shaohan Huang, Shuming Ma, Yuqing Xia, Jilong Xue, Jianyong Wang, and Furu Wei.
\newblock Retentive network: A successor to transformer for large language models.
\newblock \emph{arXiv preprint arXiv:2307.08621}, 2023.

\bibitem[Touvron et~al.(2022)Touvron, Cord, and J{\'e}gou]{touvron2022deit}
Hugo Touvron, Matthieu Cord, and Herv{\'e} J{\'e}gou.
\newblock Deit iii: Revenge of the vit.
\newblock In \emph{European conference on computer vision}, pp.\  516--533. Springer, 2022.

\bibitem[Valipour et~al.(2023)Valipour, Rezagholizadeh, Rajabzadeh, Tahaei, Chen, and Ghodsi]{valipour2023sortednet}
Mojtaba Valipour, Mehdi Rezagholizadeh, Hossein Rajabzadeh, Marzieh Tahaei, Boxing Chen, and Ali Ghodsi.
\newblock Sortednet, a place for every network and every network in its place: Towards a generalized solution for training many-in-one neural networks.
\newblock \emph{arXiv preprint arXiv:2309.00255}, 2023.

\bibitem[Vaswani et~al.(2017)Vaswani, Shazeer, Parmar, Uszkoreit, Jones, Gomez, Kaiser, and Polosukhin]{vaswani2017attention}
Ashish Vaswani, Noam Shazeer, Niki Parmar, Jakob Uszkoreit, Llion Jones, Aidan~N Gomez, {\L}ukasz Kaiser, and Illia Polosukhin.
\newblock Attention is all you need.
\newblock \emph{Advances in neural information processing systems}, 30, 2017.

\bibitem[Waleffe et~al.(2024)Waleffe, Byeon, Riach, Norick, Korthikanti, Dao, Gu, Hatamizadeh, Singh, Narayanan, et~al.]{waleffe2024empirical}
Roger Waleffe, Wonmin Byeon, Duncan Riach, Brandon Norick, Vijay Korthikanti, Tri Dao, Albert Gu, Ali Hatamizadeh, Sudhakar Singh, Deepak Narayanan, et~al.
\newblock An empirical study of mamba-based language models.
\newblock \emph{arXiv preprint arXiv:2406.07887}, 2024.

\bibitem[Yu \& Huang(2019)Yu and Huang]{yu2019universally}
Jiahui Yu and Thomas~S Huang.
\newblock Universally slimmable networks and improved training techniques.
\newblock In \emph{Proceedings of the IEEE/CVF international conference on computer vision}, pp.\  1803--1811, 2019.

\bibitem[Yu et~al.(2018)Yu, Yang, Xu, Yang, and Huang]{yu2018slimmable}
Jiahui Yu, Linjie Yang, Ning Xu, Jianchao Yang, and Thomas Huang.
\newblock Slimmable neural networks.
\newblock \emph{arXiv preprint arXiv:1812.08928}, 2018.

\bibitem[Yun et~al.(2019)Yun, Han, Oh, Chun, Choe, and Yoo]{yun2019cutmix}
Sangdoo Yun, Dongyoon Han, Seong~Joon Oh, Sanghyuk Chun, Junsuk Choe, and Youngjoon Yoo.
\newblock Cutmix: Regularization strategy to train strong classifiers with localizable features.
\newblock In \emph{Proceedings of the IEEE/CVF international conference on computer vision}, pp.\  6023--6032, 2019.

\bibitem[Zhang \& Sennrich(2019)Zhang and Sennrich]{zhang2019root}
Biao Zhang and Rico Sennrich.
\newblock Root mean square layer normalization.
\newblock \emph{Advances in Neural Information Processing Systems}, 32, 2019.

\bibitem[Zhang(2017)]{zhang2017mixup}
Hongyi Zhang.
\newblock mixup: Beyond empirical risk minimization.
\newblock \emph{arXiv preprint arXiv:1710.09412}, 2017.

\bibitem[Zhong et~al.(2020)Zhong, Zheng, Kang, Li, and Yang]{zhong2020random}
Zhun Zhong, Liang Zheng, Guoliang Kang, Shaozi Li, and Yi~Yang.
\newblock Random erasing data augmentation.
\newblock In \emph{Proceedings of the AAAI conference on artificial intelligence}, volume~34, pp.\  13001--13008, 2020.

\bibitem[Zhu et~al.(2024)Zhu, Liao, Zhang, Wang, Liu, and Wang]{zhu2024vision}
Lianghui Zhu, Bencheng Liao, Qian Zhang, Xinlong Wang, Wenyu Liu, and Xinggang Wang.
\newblock Vision mamba: Efficient visual representation learning with bidirectional state space model.
\newblock \emph{arXiv preprint arXiv:2401.09417}, 2024.

\end{thebibliography}
\bibliographystyle{iclr2025_conference}

\newpage
\appendix
\section{Appendix\label{sec:A}}
\begin{listing}[h!]
\caption{Pytorch-style pseudocode for a MatMamba block}
\label{pseudocode}
\begin{minted}[
  numbersep=5pt,
  frame=single,
  fontsize=\tiny,
  % fontseries=md,  % Remove italics (medium series)
  baselinestretch=1.0,  % Decrease line spacing
]{python}
# Example MatMamba parameters
d_model = 1024
expand = 2
headdim = 64
d_state = 128
d_inner = expand * d_model
n_heads = d_inner // headdim

# Learnable parameters, their shapes:
w_z       # (d_inner, d_model)
w_x       # (d_inner, d_model)
w_B       # (d_state, d_model)
w_C       # (d_state, d_model)
w_dt      # (n_heads, d_model)
D         # (n_heads)
A         # (n_heads)
w_conv_x  # (d_inner, 1, 4)
w_conv_BC # (2 * d_state, 1, 4)
w_out     # (d_model, d_inner)

def matmamba_layer(x_in, mat_dims):
    '''
    Arguments:
        x_in: (batch, seq_len, d_model)
        mat_dims: how many matryoshka dims to select in this block
    Returns:
        y: (batch, seq_len, d_model)
    '''
    mat_d_inner = expand * mat_dims
    mat_n_heads = mat_d_inner // headdim
    assert mat_d_inner % headdim == 0

    # Matryoshka structure on dims of W_z and W_x, and number of heads of W_dt
    w_in_proj = torch.cat(
        [w_z[:mat_d_inner, :], w_x[:mat_d_inner, :], w_B, w_C, w_dt[:mat_n_heads, :]],
        dim=0
    )

    zxbcdt = F.linear(x_in, w_in_proj)
    z, xBC, dt = torch.split(zxbcdt, [mat_d_inner, mat_d_inner + 2*d_state, mat_n_heads], dim=-1)

    # Matryoshka structure on W_conv_x based on mat_dims
    w_conv = torch.cat([w_conv_x[:mat_d_inner], w_conv_BC])
    xBC = F.conv1d(xBC, w_conv, groups=mat_d_inner + 2 * d_state)
    x, B, C = torch.split(xBC, [mat_d_inner, d_state, d_state], dim=-1)

    # Matryoshka structure on number of heads in dt, A, and D
    y = mamba_chunk_scan_combined(x, dt[:mat_n_heads], A[:mat_n_heads], B, C, D[:mat_n_heads])

    y = rmsnorm(y * F.silu(z), w_norm)

    # Matryoshka structure on dims of W_out
    y = F.linear(y, w_out_proj[:, :mat_d_inner])

    return y
\end{minted}

\end{listing}

\begin{table}
    \centering
    \caption{Training Configuration for ImageNet runs}
    \label{tab:imagenet_training_params}
    \begin{tabular}{lll}
        \toprule
         & \multicolumn{2}{c}{MatMamba-Vision}\\
        Procedure &  135M & 35M\\
        \midrule
        Model Dim. & 1024 & 512\\
        Layers & 20 & 20\\
        Batch Size & 4096 & 8192\\
        Training Steps & 249,600 & 124,800\\
        Optimizer & AdamW & AdamW\\
        LR & 0.005 & 0.005\\
        LR Decay & Cosine & Cosine\\
        Weight decay & 0.1 & 0.1\\
        Warmup steps & 10,000 & 10,000\\
        Label smoothing eps. & 0.1 & 0.1\\
        Dropout & 0.1 & 0.1\\
        Stochastic depth & 0.1 & 0.1\\
        Repeated Aug & Yes & Yes\\
        Gradient clip & 1.0 & 1.0\\
        Horizontal flip & Yes & Yes\\
        Random Resized Crop & Yes & Yes\\
        RandAugment & (2,9) & (2,9)\\
        MixUp Alpha & 0.8 & 0.8\\
        CutMix Alpha & 1.0 & 1.0\\
        RandomErase prob. & 0.3 & 0.3\\
        ColorJitter & 0.3 & 0.3\\
        Test crop ratio & 1.0 & 1.0\\
        \bottomrule
    \end{tabular}
\end{table}



\begin{table}
    \centering
    \caption{Learnable parameters (without biases) in a MatMamba layer, with example parameter reduction from a Mamba2 layer for $d_{model}=1024$, $d_{head}=32$, $d_{inner}= 2 \times 1024 = 2048$ (expand factor 2), $d_{state}=128$, $m_{i}=512$, and $h_{i}=16$ (half of the original dimensions and half of original heads being used inside the model).}
    \label{tab:matmamba_params}
    \begin{tabular}{cccc}
        \toprule
        Parameter & Mamba Shape & MatMamba Shape & Reduction Fraction\\ 
        \midrule
        \multirow{3}{*}{$W_z$} & $d_{inner} \times d_{model}$ & $(2 \times m_i) \times d_{model}$ & \multirow{3}{*}{0.5x}\\
                              & $2048 \times 1024$           & $(2 \times 512) \times 1024$     & \\
                              & 2,097,152  & 1,048,576  & \\
        \multirow{3}{*}{$W_x$} & $d_{inner} \times d_{model}$ & $(2 \times m_i) \times d_{model}$ & \multirow{3}{*}{0.5x}\\
                              & $2048 \times 1024$           & $(2 \times 512) \times 1024$     & \\
                              & 2,097,152  & 1,048,576  & \\
        \multirow{3}{*}{$W_B$} & $d_{state} \times d_{model}$ & $d_{state} \times d_{model}$ & \multirow{3}{*}{1x}\\
                              & $128 \times 1024$            & $128 \times 1024$     & \\
                              & 131,072  & 131,072 & \\
        \multirow{3}{*}{$W_C$} & $d_{state} \times d_{model}$ & $d_{state} \times d_{model}$ & \multirow{3}{*}{1x}\\
                              & $128 \times 1024$            & $128 \times 1024$     & \\
                              & 131,072  & 131,072 & \\
        \multirow{3}{*}{$W_{dt}$} & $n_{heads} \times d_{model}$ & $h_{i} \times d_{model}$ & \multirow{3}{*}{0.5x}\\
                                 & $32 \times 1024$            & $16 \times 512$       & \\
                             & 32,768   & 16,384 & \\
        \multirow{2}{*}{$D$} & $n_{heads}$                 & $h_{i}$           & \multirow{2}{*}{0.5x}\\
                              & $32$                        & $16$                 & \\
        \multirow{2}{*}{$A$} & $n_{heads}$                 & $h_{i}$           & \multirow{2}{*}{0.5x}\\
                              & $32$                        & $16$                 & \\
        \multirow{3}{*}{$W_{conv_x}$} & $d_{inner} \times 1 \times 4$ & $2 \times m_i \times 1 \times 4$ & \multirow{3}{*}{0.5x}\\
                                      & $2048 \times 1 \times 4$      & $(2 \times 512) \times 1 \times 4$ & \\
                            & 8,192   & 4,096 & \\
        \multirow{3}{*}{$W_{conv_{BC}}$} & $(2 \times d_{state}) \times 1 \times 4$ & $(2 \times d_{state}) \times 1 \times 4$ & \multirow{3}{*}{1x}\\
                                       & $(2 \times 128) \times 1 \times 4$      & $(2 \times 128) \times 1 \times 4$   & \\
                            & 256  & 256 & \\
        \multirow{3}{*}{$W_{out}$} & $d_{model} \times d_{inner}$ & $d_{model} \times (2 \times m_i)$ & \multirow{3}{*}{0.5x}\\
                                   & $1024 \times 2048$          & $1024 \times (2 \times 512$) & \\
                            & 2,097,152  & 1,048,576 & \\
        \midrule
        Total & 6,594,880 & 3,428,640 & 0.519x\\
        \bottomrule
    \end{tabular}
\end{table}

\end{document}